\documentclass[11pt]{article}
\usepackage[margin=1in]{geometry}
\usepackage[T1]{fontenc}
\usepackage[utf8]{inputenc}
\usepackage{lmodern}
\usepackage{microtype}
\usepackage{parskip}
\usepackage{amsmath,amssymb,amsthm,mathtools}
\usepackage{graphicx}
\usepackage{xcolor}
\usepackage{bbm}
\usepackage{booktabs}
\usepackage{siunitx}
\usepackage{caption}
\usepackage{subcaption}
\usepackage[round]{natbib}
\usepackage{hyperref}
\usepackage[nameinlink,noabbrev]{cleveref}
\hypersetup{
  colorlinks=true,
  linkcolor=blue,
  citecolor=blue,
  urlcolor=blue
}
\usepackage{abstract}

\usepackage[most]{tcolorbox}
\newtcolorbox{absbox}{
  colback=white,
  colframe=black!15,
  boxrule=0.6pt,
  arc=2mm,
  left=6pt,right=6pt,top=6pt,bottom=6pt
}
\usepackage{titlesec}
\titlespacing*{\section}{0pt}{1.5ex plus .2ex}{.8ex}
\titlespacing*{\subsection}{0pt}{1.25ex plus .2ex}{.6ex}
\theoremstyle{plain}

\theoremstyle{definition}

\usepackage{algorithm}
\usepackage{dsfont}
\usepackage{algpseudocode}
\usepackage{fancyhdr}
\newcommand{\keywords}[1]{\par\smallskip\noindent\textbf{Keywords—} #1}

\fancypagestyle{firstpage}{
  \fancyhf{}
  \fancyhead[L]{%
    \parbox[b]{\textwidth}{%
      \vspace*{2pt}%
      \makebox[\textwidth][l]{\includegraphics[height=20pt]{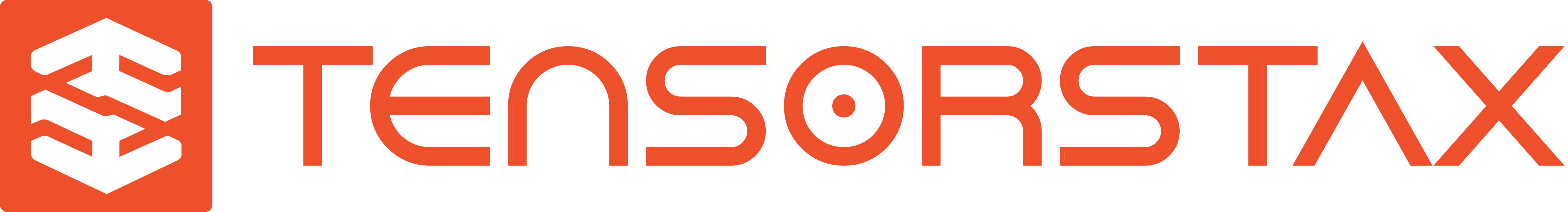}}%
      \par\vspace{-4pt}%
      \color{black!60}\rule{\textwidth}{0.4pt}%
    }%
  }
  \fancyhead[R]{%
    \raisebox{12pt}{\rmfamily October~2025}%
  }
  
}

\makeatother

\title{Thinkquel: A Model Dedicated to Text-to-dbt Using Synthetic Data and a Span-Aware Objective}
\author{
  \textbf{Anni Li$^{*}$} \\
  TensorStax \\
  \texttt{anni@tensorstax.com}
  \and
  \textbf{Aria Attar$^{*}$} \\
  TensorStax \\
  \texttt{aria@tensorstax.com}
  \and
  \textbf{Paul Dong$^{*}$} \\
  TensorStax \\
  \texttt{paul@tensorstax.com
  }
}

\begin{document}
\maketitle
\thispagestyle{firstpage}
\begingroup
\renewcommand\thefootnote{}\footnotetext{${}^*$Equal contribution}\addtocounter{footnote}{-1}
\endgroup

\begin{abstract}
\begin{absbox}
Transforming natural-language requests into reliable, production-ready data transformations remains challenging: correctness depends on precise schema linking and warehouse-specific SQL dialects, while the strongest supervision available during training—execution success and result matching—are provided only at the sequence level. At the same time, assembling large, execution-validated corpora is costly, and token-level objectives misalign with these global signals, yielding unstable optimization and limited portability. We introduce Thinkquel, a fine-tuned model for producing robust, portable, and execution-validated database queries. Methodologies in Thinkquel integrates a novel synthetic data pipeline, TS-SQL, that leverages dbt as a portable intermediate representation with a span-aware reinforcement learning objective, and Token–Sequence GRPO (TS–GRPO), specifically designed to bridge the gap between token-level training signals and sequence-level execution rewards when finetuning LLMs. On the 500-example TS–SQL test set, Thinkquel (32B) reaches 93.2\% execution success and 61.8\% exact-result match with a two-stage SFT curriculum, improving over the base model by +67.2\% (exec.) and +44.4\% (match). In Spider (14B) experiments, TS–GRPO increases training stability and speeds convergence of the execution-match reward relative to GRPO and GSPO.

\end{absbox}
\end{abstract}


\section{Introduction}
The application of Large Language Models (LLMs) in generating Structured Query Language (SQL) from natural language enables users to query complex databases without specialized expertise~\citep{li2014constructing}. However, this task presents unique and formidable challenges that distinguish it from general-purpose code generation. Unlike more permissive languages, SQL's correctness is notoriously brittle, hinging on precise schema linking~\citep{guo2019towards}, strict dialect-specific syntax, and query-level semantics where small token-level errors frequently lead to complete execution failure~\citep{zhong2017seq2sql}.

\paragraph{Why this is hard.}
Text-to-SQL differs fundamentally from code generation in languages like Python:
\begin{enumerate}
  \item \textbf{Tighter determinism, fewer degrees of freedom.} For a given intent, there are fewer acceptable SQL realizations than plausible code implementations. Many surface edits are not semantics-preserving (\texttt{INNER} vs.\ \texttt{LEFT} joins; \texttt{COUNT(*)} vs.\ \texttt{COUNT(col)} under \texttt{NULL}s; \texttt{DISTINCT} vs.\ \texttt{GROUP BY}), yielding brittle pass/fail signals~\citep{zhong2017seq2sql}.
  \item \textbf{Schema grounding and latent context.} Correctness hinges on precise linking to tables, columns, join keys, data types, time semantics, and window frames that are often implicit in the request; small grounding errors produce syntactically valid but semantically wrong queries~\citep{yu2018spider}.
  \item \textbf{Dialect dependency.} SQL dialects differ significantly in key areas, including identifier quoting, collations, result pagination, functions, and support for complex types. A query that executes on Snowflake may fail or change semantics on Postgres/BigQuery. Robust systems must either learn dialect-specific patterns or target a portable intermediate representation~\citep{ scholak2021picard}.
  \item \textbf{Scarcity of Training Data.} Most publicly available datasets consist of simple, single-shot SQL queries on limited database schemas. Creating high-quality training data that includes execution feedback is expensive~\citep{wang2021learningexecutionssemanticparsing}. As a result, there is a lack of data covering the complex, multi-step analytical patterns found in real-world use.
\end{enumerate}

\paragraph{Thinkquel: A Model Dedicated to Text-to-dbt Using Synthetic Data and a Span-Aware Objective.}

Our approach tackles the challenges of text-to-SQL generation through two primary contributions: a scalable, diverse data generation pipeline and a tailored training objective. First, to overcome the scarcity of high-quality supervision, we developed a rigorous \textbf{synthetic data pipeline} that programmatically generates, then intelligently refines and curates diverse dbt models with natural language request pairs. This ensures a steady supply of portable, execution-verified high-quality training data.

We target \textbf{dbt (Data Build Tool)} instead of raw SQL to bridge the gap between natural language requests and production-ready data transformations. While raw SQL is powerful, it lacks portability across different data warehouses and offers no built-in support for testing, documentation, or dependency management~\citep{DoorDashTransaxle}. dbt addresses these limitations by acting as a modern abstraction layer over SQL. It handles cross-dialect compilation, allows for modular and reusable code, and integrates natively with version control and CI/CD workflows~\citep{DBT}. By generating dbt models, our framework produces outputs that are not just correct, but also robust, maintainable, and immediately deployable in a modern data stack.

Second, we introduce \textbf{Token–Sequence GRPO (TS–GRPO)}, a novel, span-aware training objective specifically suitable for \textbf{reasoning text-to-SQL models}. TS–GRPO uses stable, \emph{token-level} optimizations for the concise reasoning plan, while applying more robust, length-normalized \emph{sequence-level} optimizations for the final dbt model code. This dual approach aligns the training process with the pass/fail nature of execution-based rewards, reducing optimization variance and improving model stability.

\paragraph{Contributions.}
Our contributions in this paper can be summarized as follows:
\begin{itemize}
  \item \textbf{Synthetic, portable data pipeline.} A pipeline that generates, refines, executes, and curates NL+dbt pairs, providing a reliable approach for high-quality NL-dbt data synthesis (Sec.~\ref{sec:ts-sql}).
  \item \textbf{TS–GRPO.} A span-aware training objective combining group-relative advantages~\citep{shao2024deepseekmath} with \emph{sequence-level} optimizations~\citep{zheng2025group} for answer/SQL spans and \emph{token-level} optimizations for reasoning spans, with distinct clip ranges and scalable, separate advantage and loss calculation (Sec.~\ref{sec:ts-grpo}).
  \item \textbf{Training recipe.} A "plan-before-SQL" formatting, a rewards design for stable reinforcement learning, and parameter configurations that mitigate inefficient training for text-to-dbt tasks. (Sec.~\ref{sec:training}).
\end{itemize}

\paragraph{Results in brief.}
Across text-to-SQL and text-to-DBT settings, TS–GRPO delivers (i) faster and steadier training than GRPO/GSPO on Spider (14B), (ii) state-of-the-art match within the $\leq$ 32B class on TS–SQL (32B: 93.2\% exec, 61.8\% match.), improving over GSPO by +3.6 pp match at similar or higher execution, and (iii) parity on out-of-domain BIRD–dbt (32B: 73.5\% match at 92.9\% exec.). The two-stage SFT with explicit planning supplies most of the jump from base capability to robust dbt generation; TS–GRPO then tightens execution-aligned optimization to close the remaining gap. 

\paragraph{Roadmap.}
 Sec.~\ref{sec:ts-sql} details the TensorStax-SQL (TS-SQL) synthetic pipeline, including dbt model generation, execution validation, semantic refinement, question generation and curation. Sec.~\ref{sec:training} describes our training recipe—plan-before-SQL formatting, a two-stage SFT curriculum, and reward shaping aligned with planning and execution. Sec.~\ref{sec:ts-grpo} presents Token–Sequence GRPO (TS–GRPO), defining the span decomposition, span-wise advantages, mixed token/sequence importance ratios, and separate asymmetric clipping. Sec.~\ref{sec:experiments} reports results on Spider dataset and our TS-SQL dbt dataset. The Appendix provides additional implementation details.

\section{Preliminaries}
\subsection{Synthetic Dataset}
Despite rapid progress in LLMs, high-quality datasets for text-to-query tasks (text-to-SQL, text-to-dbt) remain scarce and expensive to annotate. This gap has motivated synthetic pipelines that use programmatic schema/query generation combined with LLMs to create natural language questions. Recent systems such as Omni-SQL (SynSQL-2.5M)~\citep{li2025omnisql} demonstrate the promise of this approach: large-scale synthetic corpora generated from thousands of synthetic schemas enable training at a scale previously unattainable with human-only annotation. LLMs have transformed these pipelines by (i) generating fluent, diverse natural-language paraphrases of SQL queries, and (ii) serving as curators that filter low-quality pairs. 

\subsection{Group-Relative Policy Optimization (GRPO)}
\label{sec:prelim-grpo}
GRPO~\citep{shao2024deepseekmath} is a proximal policy optimization method that replaces an explicit critic with \emph{group-relative} advantages computed from multiple rollouts under the same context.

\paragraph{Group-relative advantage.}
Given a context $q$, sample a group of $G$ responses $\{o_i\}_{i=1}^G$ from the behavior policy $\pi_{\theta_{\mathrm{old}}}$; denote the scalar score or reward for response $o_i$ as $R_i$. The \emph{relative} advantage of a response within the group is computed as:
\[
\tilde A_i = \frac{R_i - \text{mean}(\{R_1, R_2, \dots, R_G\})}{\text{std}(\{R_1, R_2, \dots, R_G\})}
\]
which is broadcasted token-wise for the advantage of token at position $t$ in response $i$:
\begin{equation}
\hat A_{i,t}\;=\;\tilde A_i
\label{eq:grpo-adv}
\end{equation}

\paragraph{Importance ratios.}
Define the token-level importance ratio between the target policy $\pi_\theta$ and the behavior policy $\pi_{\theta_{\mathrm{old}}}$:
\begin{equation}
r_{i,t}(\theta) \;=\; 
\frac{\pi_\theta\!\left(o_{i,t}\mid q,\,o_{i,<t}\right)}
     {\pi_{\theta_{\mathrm{old}}}\!\left(o_{i,t}\mid q,\,o_{i,<t}\right)}
\label{eq:grpo-ratio}
\end{equation}
This ratio corrects the distribution mismatch when optimizing with samples drawn from $\pi_{\theta_{\mathrm{old}}}$ and is the central control knob for proximal updates.

\paragraph{Per-token proximal surrogate.}
 let $|o_i|$ be the token length and $o_{i,<t}$ the prefix up to position $t{-}1$. Following the per-token PPO-style surrogate with clipping, GRPO optimizes:
\begin{equation}
\mathcal{J}_{\text{GRPO}}(\theta)
=\mathbb{E}\!\left[
\frac{1}{G}\sum_{i=1}^{G}\;\frac{1}{|o_i|}\sum_{t=1}^{|o_i|}
\min\!\Big(\,r_{i,t}(\theta)\,\hat A_{i,t},\; 
\operatorname{clip}\!\big(r_{i,t}(\theta),\,1-\epsilon,\,1+\epsilon\big)\,\hat A_{i,t}\Big)
\;-\;\beta\,D_{\mathrm{KL}}\!\big(\pi_\theta\,\|\,\pi_{\mathrm{ref}}\big)
\right]\!
\label{eq:grpo-clip}
\end{equation}
where $\epsilon$ is the clipping parameter which in practice can be separated as $\epsilon_{low}$ and $\epsilon_{high}$,  $\beta\!\ge\!0$ scales the KL regularizer against a fixed reference policy $\pi_{\mathrm{ref}}$, and the factor $1/|o_i|$ enforces per-token normalization for sequence-length invariance.

\subsection{Group Sequence Policy Optimization (GSPO)}
\label{sec:prelim-gspo}

GSPO~\citep{zheng2025group} replaces token-level importance ratio with \emph{sequence-level} off-policy correction and clipping, aligning the optimization unit with the reward unit (entire responses). Let $q$ be a query, and let the old (behavior) and current (target) policies be $\pi_{\theta_{\mathrm{old}}}$ and $\pi_\theta$. For each $q$, sample a group of $G$ responses $\{o_i\}_{i=1}^G\sim\pi_{\theta_{\mathrm{old}}}(\cdot\mid q)$ with scalar rewards $\{R_i\}_{i=1}^G$. As in Sec.~\ref{sec:prelim-grpo}, define the normalized group-relative advantage as
\[
\tilde A_i = \frac{R_i - \text{mean}(\{R_1, R_2, \dots, R_G\})}{\text{std}(\{R_1, R_2, \dots, R_G\})},\qquad
{\hat A_{i,t}\;=\;\tilde A_i}
\]

\paragraph{Sequence-level importance ratio.}
Instead of token-level importance ratio, GSPO defines a \emph{length-normalized} sequence-level importance ratio
\begin{equation}
s_i(\theta)
\;=\;
\exp\!\Bigg(
\frac{1}{|o_i|}\sum_{t=1}^{|o_i|}\log\frac{\pi_\theta\!\left(o_{i,t}\mid q,\,o_{i,<t}\right)}
{\pi_{\theta_{\mathrm{old}}}\!\left(o_{i,t}\mid q,\,o_{i,<t}\right)}
\Bigg)
\;=\;
\Bigg(
\frac{\pi_\theta(o_i\mid q)}{\pi_{\theta_{\mathrm{old}}}(o_i\mid q)}
\Bigg)^{\!\!1/|o_i|}
\label{eq:gspo-ratio}
\end{equation}

This geometric-mean normalization controls variance and yields a clip range that is comparable across different sequence lengths.

\paragraph{Sequence-level proximal surrogate.}
GSPO applies clipping and optimization at the \emph{sequence} level rather than per token. With expectation over $q$ and groups sampled from $\pi_{\theta_{\mathrm{old}}}$, the clipped surrogate is
\begin{equation}
\mathcal{J}_{\mathrm{GSPO}}(\theta)
\;=\;
\mathbb{E}\!\left[
\frac{1}{G}\sum_{i=1}^{G}
\min\!\Big(s_i(\theta)\,\tilde A_i,\;
\operatorname{clip}\!\big(s_i(\theta),\,1-\epsilon,\,1+\epsilon\big)\,\tilde A_i\Big)
\right]
\label{eq:gspo-obj}
\end{equation}

which can be optionally combined with a KL regularizer to a reference policy~\citep{ouyang2022training,schulman2017proximal}. Differentiating \eqref{eq:gspo-obj} shows that gradients distribute \emph{uniformly over tokens} within each sequence (modulo the clip gate), via the factor $\frac{1}{|o_i|}\sum_{t=1}^{|o_i|}\nabla_\theta\log\pi_\theta(o_{i,t}\mid q,o_{i,<t})$ induced by \eqref{eq:gspo-ratio}, in line with sequence-level policy-gradient formulations~\citep{rennie2017self}. This avoids the token-level variance amplification seen when weighting each position by its own importance ratio~\citep{zheng2025group}.

Compared to GRPO, GSPO (i) matches reward and optimization granularity (sequence-level), (ii) uses a length-normalized sequence likelihood ratio for more stable clipping across response lengths, and (iii) empirically improves stability and efficiency—particularly in settings where token-level ratios are volatile~\citep{zheng2025group,shao2024deepseekmath}.

\section{Dataset Pipeline: TensorStax-SQL (TS-SQL)}
\label{sec:ts-sql}

\paragraph{Schema and Coverage.}
To ground model generation in realistic relational structure and naming conventions, we use schemas from the \emph{Spider}~\citep{yu2018spider}, \emph{Spider~2.0}~\citep{lei2024spider}, and \emph{BIRD}~\citep{li2023bird} databases.\footnote{We use these corpora as schema sources only; all dbt transformations and natural-language questions are generated by our pipeline.} We split the mixed databases into non-overlapped $185$ training databases and $79$ test databases to ensure data integrity.

\paragraph{Dataset Generation.}

\begin{figure}[h!]
    \centering
    \includegraphics[width=1.0\textwidth]{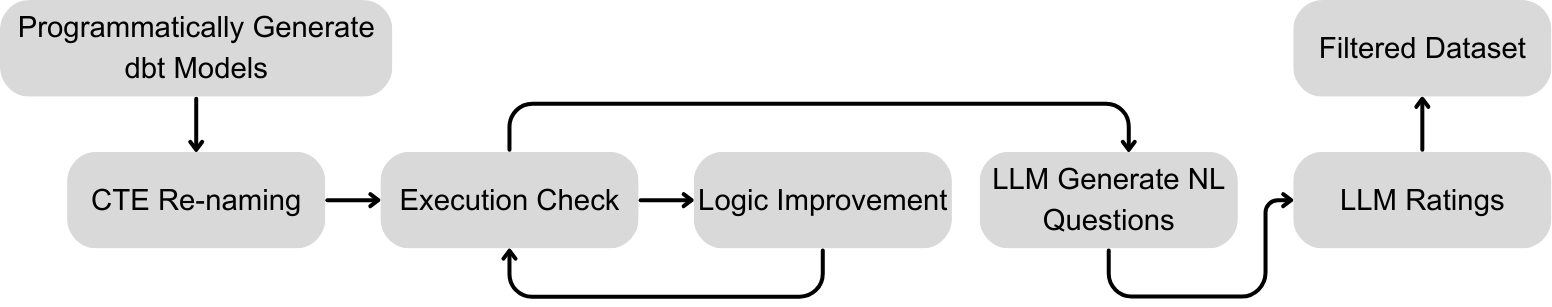}
    \caption{TensorStax-SQL Data Synthesis Pipeline.}
    \label{fig:data_pipeline}
\end{figure}

Fig.~\ref{fig:data_pipeline} illustrates TS-SQL generation pipeline. Our approach begins with a programmatic generation of millions of dbt model configurations through systematic variation of structural parameters including CTEs and different transformations, alongside SQL features such as set operations, conditional aggregates, and subqueries. Unlike template-based methods, this combinatorial approach explores a wider complexity space of analytical transformations across staging, intermediate, mart, and report model types. Each generated model then executes against its target database, with models containing syntax errors, invalid references, or timeouts got filtered out to ensure syntactic validity.

The programmatically generated models initially contain generic CTE names like \texttt{CTE1} and \texttt{CTE2}, and column names like \texttt{col1} and \texttt{col2}, which lack semantic meaning. We address this through refinement with Qwen3-Coder-480B~\citep{yang2025qwen3}, which analyzes the transformation logic and schema context to produce meaningful identifiers and at the same time enhance the overall logic. After refinement, models undergo re-execution to ensure continued validity. For each validated dbt model, we then generate diverse natural language questions using Qwen3-Coder-480B with variations in description vagueness and syntax requirements.

We use Anthropic's \textit{claude-sonnet-4-20250514} as a quality control to evaluate each question–model pair on clarity, semantic alignment, and technical correctness. Pairs scoring below 8/10 undergo targeted re-generation of either questions or models, followed by re-evaluation. Only pairs achieving scores of 9/10 or higher of both the question and the model pass the final filtering threshold. The curated dataset is then partitioned based on execution results: pairs that successfully return non-empty data feed into reinforcement learning pipelines where models learn from concrete results, while valid transformations producing empty results due to data-specific filtering mismatches support supervised fine-tuning focused on learning transformation patterns.

This pipeline advances synthetic text-to-database generation by extending beyond SQL to complex dbt transformations, combining execution validation with semantic evaluation for robust quality assurance, and enabling execution-aware training strategies that leverage both structural validity and concrete data grounding.

\paragraph{Dataset Statistics.}
We report the dataset sizes and statistics in Table~\ref{tab:ts-sql-stats}. ``Non-empty'' counts refer to examples whose execution produced at least one row. $avg. Q$ is the average question score of the filtered dataset, and $avg. M$ stands for the average model score of the filtered dataset: 
\begin{table}[h]
  \centering
  \small
  \caption{TS–SQL Split Sizes and Rating-based Filtering.}
  \label{tab:ts-sql-stats}
  \begin{tabular}{lrrrrrr}
    \toprule
    \textbf{Split} & \textbf{Original} & \textbf{Filtered} & \textbf{Non-empty} & \textbf{avg.Q} & \textbf{avg.M} \\
    \midrule
    Training & 146{,}032 & 82{,}925 & 33{,}982 & 9.39 & 9.33 \\
    Test     & 23{,}127  & 10{,}891 & 10{,}891 & 9.21 & 9.20 \\
    \bottomrule
  \end{tabular}
\end{table}

\noindent We additionally track complexity distribution (post-filtering) as shown in Table~\ref{tab:ts-sql-complexity}:
\begin{table}[h]
  \centering
  \small
  \caption{TS-SQL Complexity Distribution After Filtering.}
  \label{tab:ts-sql-complexity}
  \begin{tabular}{lrrrr}
    \toprule
    \textbf{Split} & \textbf{Simple} & \textbf{Medium} & \textbf{Complex} & \textbf{Total} \\
    \midrule
    Training & 1{,}197 (1.4\%) & 52{,}952 (63.9\%) & 28{,}776 (34.7\%) & 82{,}925 \\
    Test     & 339 (3.1\%)    & 8{,}108 (74.4\%)  & 2{,}444 (22.4\%) & 10{,}891 \\
    \bottomrule
 \end{tabular}
\end{table}

For the actual training (SFT \& RL), we only used a portion of the dataset. The details can be found in Sec.~\ref{sec:training}. For the evaluation, we randomly selected $500$ samples from the test dataset and ensured all the test databases and model types are covered.






\section{Methodology}
\subsection{Training}
\label{sec:training}

\begin{figure}[h!]
    \centering
    \includegraphics[width=1.0\textwidth]{"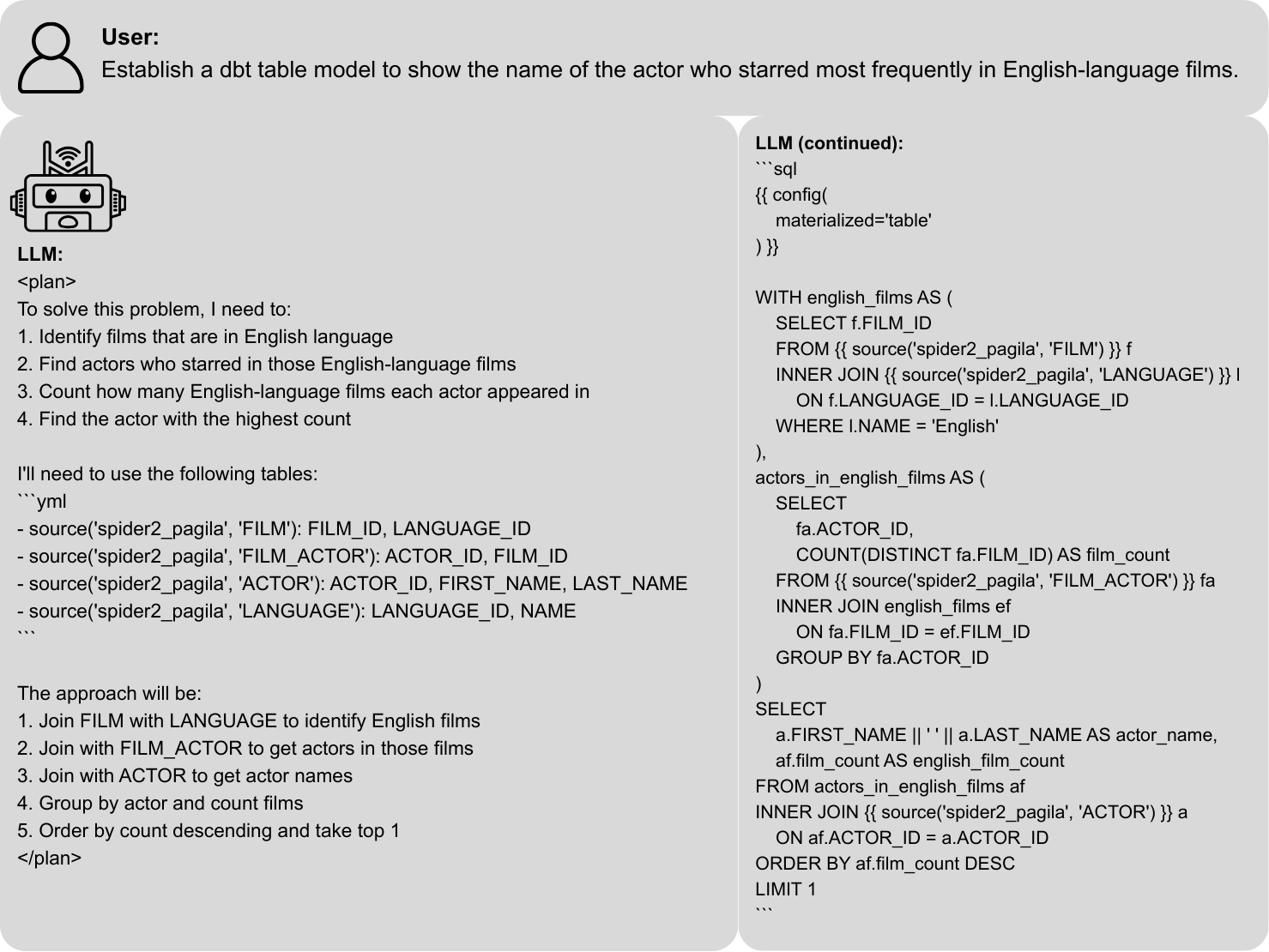"}
    \caption{An Example of the Model's Completion.}
    \label{fig:output_example}
\end{figure}

\paragraph{Concise Planning before SQL.}
Instead of verbose chain-of-thought, we train the model to first generate a concise, structured plan before the final answer, as shown in Figure~\ref{fig:output_example}. We observed that long chain-of-thought narrations often introduced noise and spurious reasoning steps, especially for smaller models. By contrast, training on \emph{concise, plan-style explanations} led to more reliable and generalizable models for several key reasons:

\begin{itemize}
    \item \textbf{Structured Scaffolding.} A plan acts as a high-level scaffold for the subsequent code~\citep{jiang2023selfplanning}. It forces the model to first identify the necessary tables and columns, define the constituent sub-problems (e.g., course metrics, student assignments), and outline the final assembly logic (e.g., `UNION ALL`) before committing to low-level syntax.
    \item \textbf{Improved Schema Grounding.} The yml section in the plan explicitly requires the model to list the source tables and columns it will use. This step improves schema grounding by focusing the model's attention on the available data entities early in the process, reducing hallucinations and incorrect table or column selections~\citep{Pourreza2025ReasoningSQL}.
    \item \textbf{Verifiable Planning Reward.} Unlike typical reasoning tasks that rely on a subjective reward model, our structured plans allow for objective rewards. We can more directly reward the plan's quality by measuring the percentage of data sources in the plan that match those in the ground-truth query.
    \item \textbf{Reduced Error Propagation.} Unlike free-form chain-of-thought, the structured nature of a plan makes it less likely for a small error in reasoning to cascade and derail the entire generation. The plan isolates distinct logical steps (e.g., CTE 1, CTE 2), making the generation process more robust.
\end{itemize}

\begin{figure}[h!]
    \centering
    \includegraphics[width=1.0\textwidth]{"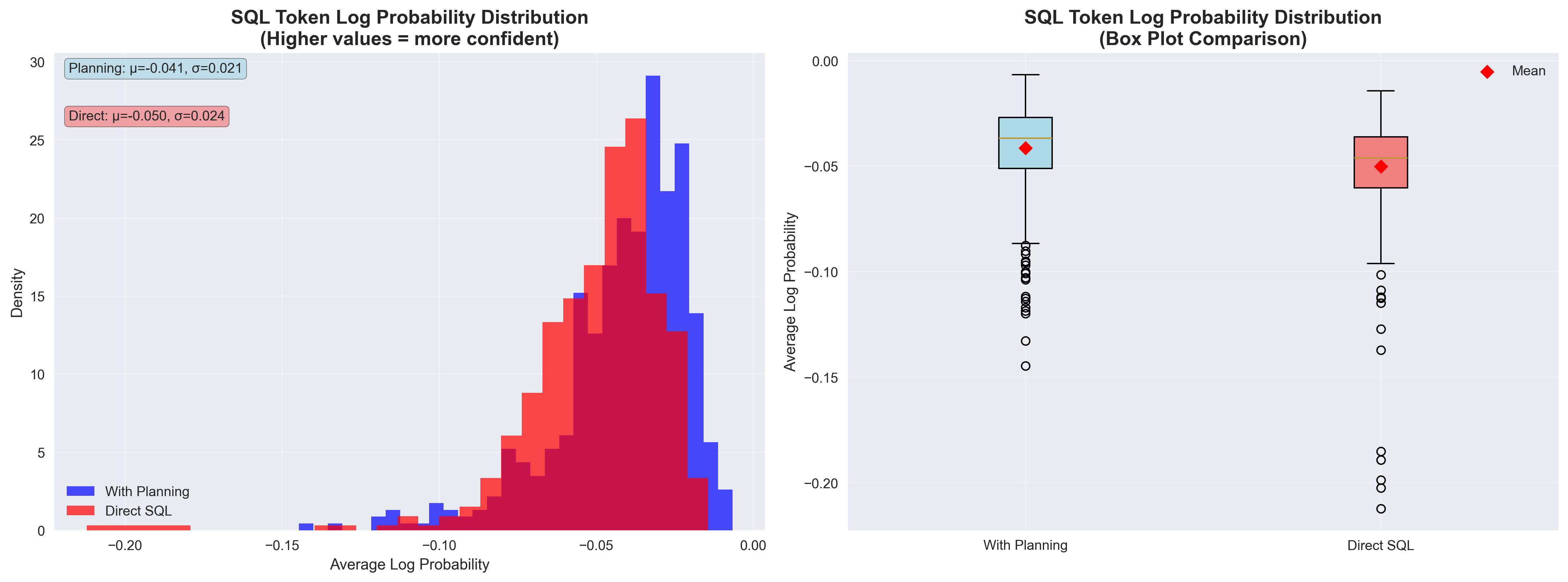"}
    \caption{Distribution of Average log-probability of SQL Token with Planning vs. No Planning}
    \label{fig:plan_logprob}
\end{figure}

To validate the necessity of planning, we measure the average SQL-token log-probability of each response using our evaluation dataset and prompt the same model to output (i) plan and SQL (ii) only the SQL. Figure~\ref{fig:plan_logprob} shows that the distribution of the log-prob shifts to the right when the model is prompted to plan, indicating the increase of confidence over SQL after planning.

\paragraph{Supervised Fine-Tuning (SFT).}
We apply a two-stage SFT curriculum prior to reinforcement learning. 
\begin{itemize}
    \item \textbf{SFT Stage 1: Base Fine-Tuning.} We fine-tune the base model on $22{,}000$ synthetic text-to-dbt pairs for one epoch to obtain the base ability for text-to-dbt tasks.
     \item \textbf{SFT Stage 2: Planning Behavior Tuning.} We further fine-tune the checkpoint from above for two epochs on a mixture of $793$ text-to-dbt instances augmented with concise \emph{planning} and $100$ general instruction-following data from Personas dataset~\citep{ge2025scalingsyntheticdatacreation} to retain broad conversational ability~\citep{wang2024demystifyinginstructionmixingfinetuning}.
\end{itemize}

\paragraph{Reinforcement Learning and Reward design.}
\label{para:reward}
We used $17,687$ samples from TS-SQL training dataset for the reinforcement learning experiments. Data used for RL is guaranteed to be executable and return non-empty results.

Our RL signal is a composite of multiple rewards that align model behavior with execution-grounded correctness while encouraging good planning and schema use:
\begin{itemize}
  \item \textbf{Format reward}: Gives credit only if the output contains both a \texttt{<plan>} with \verb|```|\texttt{yml} block and a \verb|```|\texttt{sql} block. This prevents assigning reward to malformed responses and keep the wanted model behavior.
  \item \textbf{Schema-linking (table)}: Jaccard similarity between the set of source tables listed in the \emph{plan} to the source tables used in the \emph{ground truth query}. It teaches the model to relate and name the right tables during planning.
  \item \textbf{Schema-linking (column)}: Column-level analogue of the above. It encourages precise column selection in the plan.
  \item \textbf{Plan following (table)}: Jaccard similarity between the sets of source table names actually used by the \emph{predicted dbt model} and those in the \emph{plan}. This penalizes “bait-and-switch” where the plan says one thing but the generated dbt uses another.
  \item \textbf{Plan following (column)}: Column-level analogue of the above, ensuring the SQL follows the planned column usage.
  \item  \textbf{Execution reward}: Checks whether the predicted query compiles and runs within a timeout on the target warehouse to enforces basic runnability and provide more intermediate feedbacks.
  \item \textbf{Result-match reward}: Compares the predicted query’s result to the gold result, rewards exact matches. This is the strongest signal for the learning objectives.
\end{itemize}

For span-aware RL, we combine the above itemized rewards into:
\begin{itemize}
  \item \emph{SQL-span reward} (used for the answer/dbt span): result match, execution, and plan following (tables/columns).
  \item \emph{Plan-span reward} (used for the reasoning/plan span): format, schema-linking (tables/columns).
\end{itemize}
This plan/SQL split is driven by TS–GRPO’s mixed token–sequence optimization, which we will introduce in Sec.~\ref{sec:ts-grpo}

\subsection{Algorithm: Token–Sequence GRPO (TS–GRPO)}
\label{sec:ts-grpo}
\paragraph{Motivation.}
Training text-to-dbt (and text-to-SQL) models exposes a persistent \emph{granularity mismatch}: the strongest supervision (execution \& result-match) is intrinsically \emph{sequence-level}, while standard GRPO-style updates weight \emph{every token} by its own importance ratio (Eq.~\eqref{eq:grpo-ratio}) and a group-shared advantage (Eq.~\eqref{eq:grpo-adv}). TS-GRPO is designed to meet the following desiderata:

\begin{itemize}
  \item \textbf{Isolate credit across spans.} Split total rewards and compute \emph{separate} group-relative advantages (Eqs.~\eqref{eq:ts-adv-sql}--\eqref{eq:ts-adv-rea}) for plan vs.\ SQL, then broadcast \emph{only within} each span (Eq.~\eqref{eq:ts-adv-broadcast}) to prevent plan tokens from absorbing execution credit and to prevent SQL tokens from being steered by formatting rewards.
  
  \item \textbf{Align optimization with supervision.} Use a sequence-level, length-normalized importance ratio \emph{only} on the SQL span (Eq.~\eqref{eq:ts-ratio-sql}) so that the unit of credit assignment matches sequence-level SQL-related rewards.
  \item \textbf{Preserve local structure learning.} Keep token-level ratios on the reasoning span (Eq.~\eqref{eq:ts-ratio-rea}), which is where local, schema-linking signals are available and useful.

  \item \textbf{Stabilize with length invariance.} Sequence-level updates on the SQL span backpropagate a \emph{uniform} per-token gradient (via $\nabla_\theta \log \tilde r^{\text{ans}}$) and clip in a way that is comparable across different SQL lengths, reducing variance from compounding token-wise gates.
  
  \item \textbf{Reflect asymmetric risk.} SQL correctness is brittle; planning is tolerant. TS--GRPO supports asymmetric clip ranges (Eq.~\eqref{eq:ts-clips}) with tighter SQL clips ($\epsilon_{\text{sql}}$ smaller) and looser plan clips to encourage exploration in planning while keeping SQL updates conservative.
\end{itemize}

\noindent In summary, TS--GRPO routes \emph{global, brittle} signals (execution/match/plan-following) through a length-normalized sequence update on the SQL span, while routing \emph{local, structural} signals (format/schema-linking) through token-level updates on the plan span. This span-aware routing reduces variance, curbs cross-span credit leakage, and better matches the error surface of text-to-dbt generation than uniform token- or sequence-only objectives.

\paragraph{Completion by Spans.}
TS–GRPO is a span-aware variant of GRPO that couples group-relative advantages with mixed \emph{token/sequence level} importance ratios on disjoint spans of a completion. We assume each completion $o_i$ decomposes into a \emph{reasoning} span and a \emph{answer} span. Let
\[
S_i^{\text{ans}}\subseteq\{1,\dots,|o_i|\},\qquad
S_i^{\text{rea}}\subseteq\{1,\dots,|o_i|\};\qquad
S_i^{\text{ans}}\cap S_i^{\text{rea}}=\varnothing,\qquad
S_i^{\text{ans}}\cup S_i^{\text{rea}}=\{1,\dots,|o_i|\}
\]
denote token index sets for the two spans, with lengths $|S_i^{\text{ans}}|$ and $|S_i^{\text{rea}}|$. We write $M_{i,t}^{\text{ans}}=\mathbbm{1}[t\in S_i^{\text{ans}}]$ and $M_{i,t}^{\text{rea}}=\mathbbm{1}[t\in S_i^{\text{rea}}]$ for the corresponding masks.

\paragraph{Advantage calculation.}
We split the scalar reward into span-specific components, as discussed in Sec.~\ref{para:reward}:
\[
R_i^{\text{ans}} \;\text{(execution/match/plan following)},\qquad
R_i^{\text{rea}} \;\text{(format/schema linking)}.
\]
For each query $q$, with a group of $G$ samples $\{o_i\}_{i=1}^G\sim\pi_{\theta_{\mathrm{old}}}(\cdot\mid q)$, we form group-relative (optionally std-normalized) advantages \emph{separately} for the two reward channels:
\begin{equation}
\tilde A_i^{\text{ans}} = \frac{R_i^{\text{ans}} - \text{mean}(\{R_1^{\text{ans}}, R_2^{\text{ans}}, \dots, R_G^{\text{ans}}\})}{\text{std}(\{R_1^{\text{ans}}, R_2^{\text{ans}}, \dots, R_G^{\text{ans}}\})}
\label{eq:ts-adv-sql}
\end{equation}

\begin{equation}
\tilde A_i^{\text{rea}} = \frac{R_i^{\text{rea}} - \text{mean}(\{R_1^{\text{rea}}, R_2^{\text{rea}}, \dots, R_G^{\text{rea}}\})}{\text{std}(\{R_1^{\text{rea}}, R_2^{\text{rea}}, \dots, R_G^{\text{rea}}\})}
\label{eq:ts-adv-rea}
\end{equation}

We then broadcast these scalars to each token \emph{by their spans}:
\begin{equation}
\hat A_{i,t}
=\tilde A_i^{\text{ans}}\,M_{i,t}^{\text{ans}}
+\tilde A_i^{\text{rea}}\,M_{i,t}^{\text{rea}}.
\label{eq:ts-adv-broadcast}
\end{equation}

\paragraph{Importance ratios and clipping.}
Let the per-token importance ratio be as in \eqref{eq:grpo-ratio} and per-sequence importance ratio be as in \eqref{eq:gspo-ratio}:
\[
r_{i,t}(\theta)
=\tfrac{\pi_\theta(o_{i,t}\mid q,o_{i,<t})}{\pi_{\theta_{\mathrm{old}}}(o_{i,t}\mid q,o_{i,<t})},\qquad
s_i(\theta)
\;=\;
\Bigg(
\frac{\pi_\theta(o_i\mid q)}{\pi_{\theta_{\mathrm{old}}}(o_i\mid q)}
\Bigg)^{\!\!1/|o_i|}
\]

TS–GRPO uses \emph{sequence-level} correction on the answer span and \emph{token-level} correction on the reasoning span:
\begin{align}
\tilde r_i^{\text{ans}}(\theta)
&=\exp\!\Big(\frac{1}{|S_i^{\text{ans}}|}\sum_{t\in S_i^{\text{ans}}}\log r_{i,t}(\theta)\Big)
=\Bigg(\frac{\prod_{t\in S_i^{\text{ans}}}\pi_\theta(o_{i,t}\mid q,o_{i,<t})}
             {\prod_{t\in S_i^{\text{ans}}}\pi_{\theta_{\mathrm{old}}}(o_{i,t}\mid q,o_{i,<t})}\Bigg)^{\!\!\!1/|S_i^{\text{ans}}|},
\label{eq:ts-ratio-sql}\\[4pt]
r_{i,t}^{\text{rea}}(\theta)&=r_{i,t}(\theta),\qquad t\in S_i^{\text{rea}}.
\label{eq:ts-ratio-rea}
\end{align}
We allow different clip ranges for the two spans:
\begin{equation}
\operatorname{clip}_{\text{ans}}(x)=\mathrm{clip}(x,\,1-\epsilon^{\text{low}}_{\text{ ans}},\,1+\epsilon^{\text{high}}_{\text{ans}}),\qquad
\operatorname{clip}_{\text{rea}}(x)=\mathrm{clip}(x,\,1-\epsilon^{\text{low}}_{\text{rea}},\,1+\epsilon^{\text{high}}_{\text{rea}}).
\label{eq:ts-clips}
\end{equation}

\paragraph{Gradient update.}
TS–GRPO optimizes a \emph{span-wise} clipped PPO surrogate with length-invariant normalization inside each span:
\begin{align}
\mathcal{J}_{\text{TS--GRPO}}(\theta)
&=\mathbb{E}\Bigg[
\frac{1}{G}\sum_{i=1}^{G}\Bigg\{
\underbrace{\alpha_{\text{ans}}\;
\min\!\Big(\tilde r_i^{\text{ans}}(\theta)\,\tilde A_i^{\text{ans}},\;
\operatorname{clip}_{\text{ans}}\!\big(\tilde r_i^{\text{ans}}(\theta)\big)\,\tilde A_i^{\text{ans}}\Big)}_{\text{SQL (sequence-level) term}}
\\[-2pt]
&\qquad\qquad
+\underbrace{\frac{\alpha_{\text{rea}}}{|S_i^{\text{rea}}|}\sum_{t\in S_i^{\text{rea}}}
\min\!\Big(r_{i,t}^{\text{rea}}(\theta)\,\tilde A_i^{\text{rea}},\;
\operatorname{clip}_{\text{rea}}\!\big(r_{i,t}^{\text{rea}}(\theta)\big)\,\tilde A_i^{\text{rea}}\Big)}_{\text{Reasoning (token-level) term}}
\Bigg\}
\\
&\qquad
-\;\beta_{\text{ans}}\,\mathrm{KL}_{\text{ans}}(\theta)
-\;\beta_{\text{rea}}\,\mathrm{KL}_{\text{rea}}(\theta)
\Bigg].
\label{eq:ts-obj}
\end{align}

where $\alpha_{\text{ans}},\alpha_{\text{rea}}\!\ge\!0$ balance loss contributions from different spans, and $\mathrm{KL}_{\text{ans}}$ / $\mathrm{KL}_{\text{rea}}$ are optional regularizers to a fixed reference policy restricted to the respective spans. The SQL term’s gradient distributes uniformly over its span via
\[
\nabla_\theta \log \tilde r_i^{\text{ans}}(\theta)
=\frac{1}{|S_i^{\text{ans}}|}\sum_{t\in S_i^{\text{ans}}}\nabla_\theta \log \pi_\theta(o_{i,t}\mid q,o_{i,<t}),
\]
which matches the reward's sequence granularity and stabilizes clipping across lengths~\citep{sutton1999policy, williams1992reinforce}. The reasoning term uses standard token-level reweighting.

\paragraph{Design notes.}
(i) Equations \eqref{eq:ts-adv-sql}–\eqref{eq:ts-adv-broadcast} compute two group-relative advantages \emph{per sample} and broadcast them only within their spans; the \emph{reasoning} span is updated with \emph{token-level} importance ratios, so it consumes local, structure-oriented signals (format; table/column linking). The \emph{answer/SQL} span is updated with \emph{length-normalized sequence-level} ratios, so it consumes global, program-level signals (execution; result match) plus consistency with the plan (plan-following). Routing rewards at the same granularity as the update rule reduces variance and prevents cross-span credit leakage. (ii) Using distinct clips $\epsilon_{\text{sql}}<\epsilon_{\text{rea}}$ often improves stability when the SQL span is decisive for task success. (iii) If a span is empty ($|S_i^{\text{rea}}|{=}0$ or $|S_i^{\text{sql}}|{=}0$), its term vanishes.

\section{Experiments}
\label{sec:experiments}

\subsection{TS–GRPO on Spider}
\paragraph{Dataset.}
Spider~\citep{yu2018spider} is a cross-domain text-to-SQL dataset consisting of natural-language questions paired with gold execution results over distinct relational databases. We use original $4,573$ training samples from the training dataset after splitting it into training and validation sets. We use the gold tables provided by Spider, which contains the source table names used in the ground truth~\citep{spider-github}, for the calculation of table linking rewards. We disable data shuffling to ensure the consistency of comparison.

\begin{figure}[h!]
    \centering
    \includegraphics[width=0.8\textwidth]{"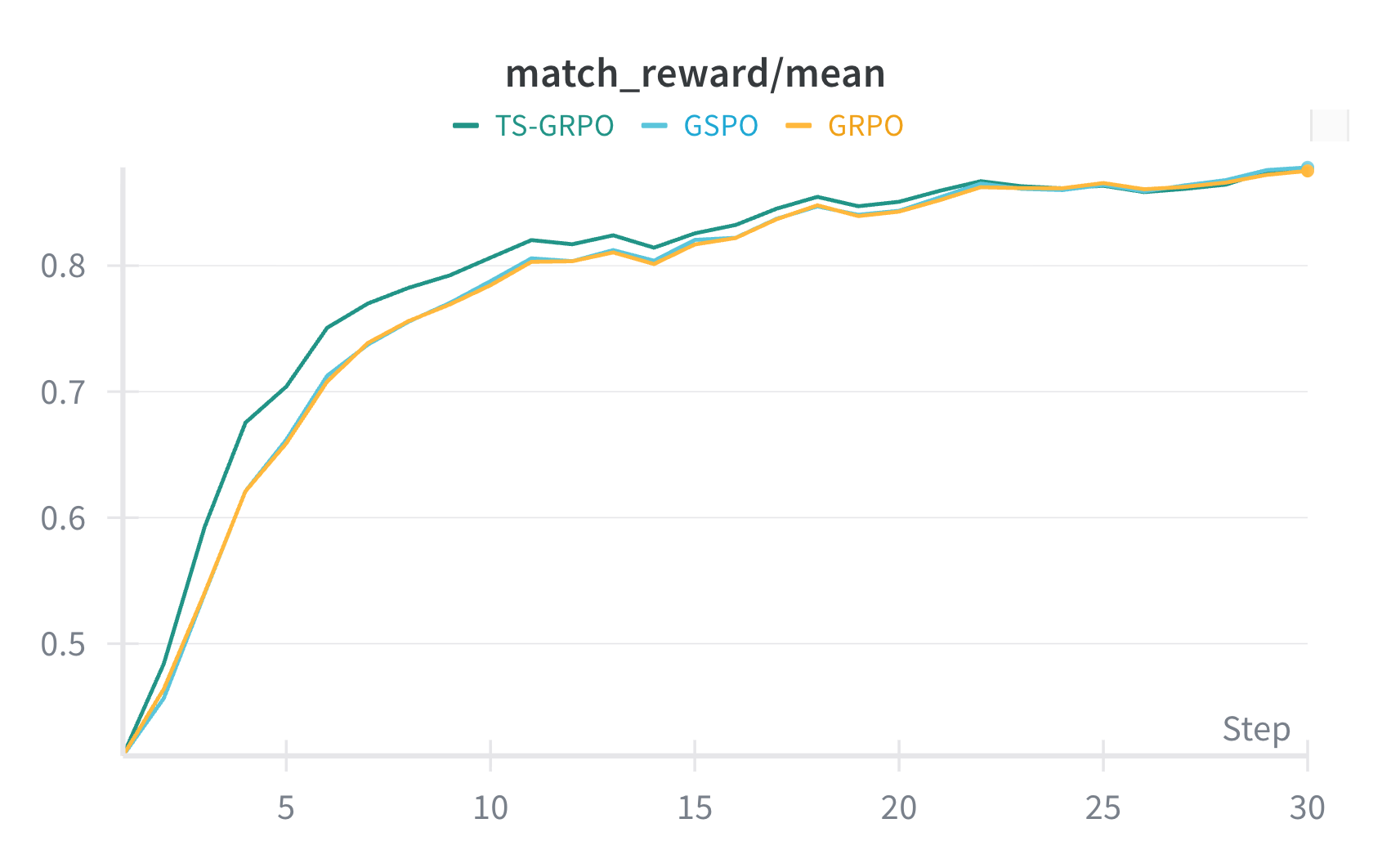"}
    \caption{TS-GRPO vs. GRPO vs. GSPO on Spider Dataset}
    \label{fig:spider1_match}
\end{figure}

\paragraph{Training.}
We compare TS–GRPO against GRPO and GSPO by fine-tuning \textbf{Qwen2.5-Coder-14B-Instruct}~\citep{qwen2024qwen25} on Spider training dataset using VeRL~\citep{verl2025} framework. To isolate the effect of the objective, we train the base model directly (no SFT) and use a prompting protocol that first elicits a concise planning span and then the final SQL enclosed in XML tags (system prompt can be found in Appendix~\ref{app:system-prompt}). Table~\ref{tab:spider_settings} lists the detailed training setting for all the experiments. Figure~\ref{fig:spider1_match} shows that TS–GRPO outpaces GSPO and GRPO in execution-match convergence on Spider while using the same backbone (Qwen2.5-Coder-14B-Instruct) and training hyperparameters (Table~\ref{tab:spider_settings}). The span-aware routing (sequence-level updates for SQL; token-level for plans) reduces variance and yields smoother learning under identical reward shaping.  

\begin{table}[h!]
\centering
\small
\begin{tabular}{l l}
\toprule
Setting & Value \\
\midrule
Model & Qwen2.5-Coder-14B-Instruct \\
Epochs & 1 \\
Batch size & 64 \\
Group size & 10 \\
Learning rate & $1{\times}10^{-6}$ \\
Temperature & 1.0\\
Hardware & 8$\times$NVIDIA A100 80GB GPUs \\
\bottomrule
\end{tabular}
\caption{Training and Rollout Settings for Spider Experiments.}
\label{tab:spider_settings}
\end{table}

\paragraph{Reward.}
Due to limitations of the Spider dataset and a lack of gold SQL, we used a different set of rewards than the experiment on TS-SQL dataset. The reward is a combination that can be expressed as
\[
R \;=\; 0.4\,R_{\text{plan}} \;+\; 0.6\,R_{\text{sql}},
\]
with \(R_{\text{plan}} = 0.2\,R_{\text{format}} + 0.8\,R_{\text{table linking}}\) and \(R_{\text{sql}} = R_{\text{match}}\), where \(R_{\text{match}}\in\{0,1\}\) indicates whether the generated query’s execution result matches the ground truth.    

\paragraph{Results.}
Figure~\ref{fig:spider1_match} plots the mean execution-match reward over training; TS–GRPO converges more rapidly than GSPO and GRPO, consistent with the intended benefit of span-targeted optimization brought by routing plan signals to token-level updates and SQL signals to sequence-level updates.

\subsection{Thinkquel on TS-SQL Dataset}
\label{sec:thinkquel-on-ts-sql}

\paragraph{Setup.}

We train the Thinkquel model using VeRL~\citep{verl2025} framework following the training pipeline introduced in Sec.~\ref{sec:training} and use \textbf{Qwen2.5\textendash Coder\textendash 32B\textendash Instruct}~\citep{qwen2024qwen25} as the base model. \emph{SFT stage 1} fine-tunes on $22{,}000$ TS-SQL samples for one epoch; \emph{SFT stage 2} refines on a mixture of $793$ plan-augmented instances (with a concise \texttt{<plan>} span) and a mixture of $100$ general instruction-following data for two epochs. For reinforcement learning, we use $17{,}687$ training examples from TS\textendash SQL that are \emph{guaranteed} to compile and return non-empty results (Sec.~\ref{para:reward}). Training configs and reward weights for Thinkquel can be found in Table~\ref{tab:thinkquel-configs} and Tabel~\ref{tab:thinkquel-weights}.

\begin{table}[h!]
\centering
\small
\begin{tabular}{l l}
\toprule
Setting & Value \\
\midrule
Epochs & 1 \\
Batch size & 32 \\
Group size & 16 \\
Learning rate & $5{\times}10^{-6}$ \\
Temperature & 1.0 \\
Top p & 0.95 \\
\bottomrule
\end{tabular}
\caption{Thinkquel Training Configs.}
\label{tab:thinkquel-configs}
\end{table}

\begin{table}[h!]
\centering
\small
\begin{tabular}{l l}
\toprule
Reward & Weight \\
\midrule
Execution & 0.2 \\
Match & 0.5 \\
Format & 0.1 \\
Schema Linking (table) & 0.07 \\
Schema Linking (column) & 0.05 \\
Plan Following (table) & 0.05 \\
Plan Following (column) & 0.03 \\
\bottomrule
\end{tabular}
\caption{Thinkquel Reward Weights.}
\label{tab:thinkquel-weights}
\end{table}

\paragraph{Evaluation.}
To get more comprehensive evaluation results, we take $1,482$ samples from the original BIRD test dataset and rewrite the questions such that they are more similar to the TS-SQL style and require the model to write dbt instead of SQL while keeping the query result the same. We call this \textit{BIRD-dbt} for short.

Results shown in Table~\ref{tab:results} compares the \textbf{execution success} (query compiles and executes within time limit) and \textbf{match accuracy}(result set perfect match) of the following models and other OSS/proprietary models evaluated with the 500 TS-SQL test dataset and the BIRD-dbt dataset:
\begin{enumerate}
    \item \textbf{Base Model}: Qwen2.5-Coder-32B-Instruct
    \item \textbf{SFT stage 1}: no RL; SFT of base model on pure question-dbt pairs.
    \item \textbf{SFT stage 2}: no RL; additional SFT of SFT-stage-1 checkpoint on question-dbt (with planning) pairs.
    \item \textbf{SFT+GSPO}: RL until converge using GSPO on the SFT checkpoint
    \item \textbf{SFT+TS-GRPO (Thinkquel)}: RL until converge using TS-GRPO on the SFT checkpoint
\end{enumerate}

\begin{table}[h!]
  \centering
  \small
  \caption{Evaluation on TS–SQL and BIRD–dbt}
  \label{tab:results}
  \begin{tabular}{lccccc}
    \toprule
    & & \multicolumn{2}{c}{\textbf{TS–SQL(500)}} & \multicolumn{2}{c}{\textbf{BIRD–dbt(1,482)}} \\
    \cmidrule(lr){3-4}\cmidrule(lr){5-6}
    \textbf{Model} & \textbf{Size} & \textbf{Exec. (\%)} & \textbf{Match (\%)} & \textbf{Exec. (\%)} & \textbf{Match (\%)} \\
    \midrule
    \multicolumn{6}{l}{\textbf{Models $\leq$ 32B}} \\
    \addlinespace[2pt]
    GPT-OSS-20B                              & 20B & \,56.8 & \,26.0 & \,76.1 & \,58.8 \\
    Qwen2.5-Coder-32B-Instruct               & 32B  & \,26.0 & \,17.4 & \,49.6 & \,38.9 \\
    SFT-stage-1                              & 32B  & \,89.0 & \,58.2 & \,79.5 & \,56.1 \\
    SFT-stage-2                              & 32B  & \,88.6 & \,53.6 & \,92   & \,70.7 \\
    SFT + GSPO                               & 32B  & \,90.4 & \,58.2 & \,\textbf{93.1} & \,73.3 \\
    SFT + TS–GRPO (Thinkquel step 100)       & 32B  & \,92.2 & \,58.5 & \,92.9 & \,\textbf{73.5} \\
    SFT + TS–GRPO (Thinkquel step 250)       & 32B  & \,\textbf{93.2} & \,\textbf{61.8} & \,92.2 & \,72.3 \\

    \midrule
    \multicolumn{6}{l}{\textbf{Models $>$ 32B or unknown}} \\
    \addlinespace[2pt]
    GPT-5-mini                               & --   & \,43.2 & \,27.4 & \,47.7  & \,38.0   \\  
    GPT-5 (thinking-high)                    & --   & \,\textbf{84.8} & \,39.2 & \,\textbf{98.0} & \,\textbf{76.5} \\    
    GPT-OSS-120B                             & 120B & \,49.4 & \,27.4 & \,70.9 & \,55.2 \\
    Claude Sonnet 4                          & --   & \,58.8 & \,32.0 & \,80.8 & \,60.9 \\
    Claude Opus 4.1 (extended thinking)      & --   & \,75.2 & \,\textbf{41.1} & \,95.8 & \,75.8 \\
    Qwen3-Coder-480B-A35B-Instruct           & 480B & \,62.8 & \,30.2 & \,87.2 & \,67.2 \\
    \bottomrule
  \end{tabular}
\end{table}

\begin{figure}[h!]
    \centering
    \includegraphics[width=0.8\textwidth]{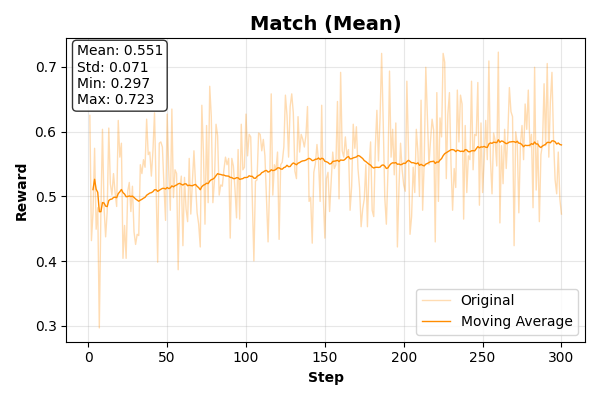}
    \caption{Learning Curves of Thinkquel on TS-SQL: Mean Match Reward vs. Steps.}
    \label{fig:thinkquel_match}
\end{figure}

\begin{figure}[h!]
    \centering
    \includegraphics[width=0.8\textwidth]{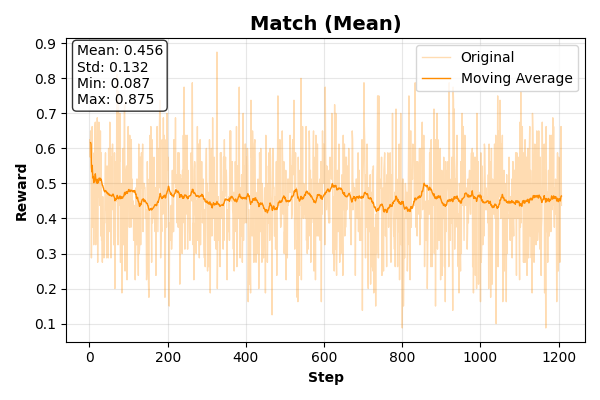}
    \caption{Learning Curves of GSPO on TS-SQL: Mean Match Reward vs. Steps.}
    \label{fig:gspo_match}
\end{figure}

\paragraph{Training dynamics.}
Figure~\ref{fig:thinkquel_match} plots Thinkquel's mean match reward over training steps, as a comparison to Figure~\ref{fig:gspo_match} which plots the mean match reward over training steps using GSPO. Both experiments were kept running until all rewards converge. We observe that for the difficult text-to-dbt setting, GSPO converges more slowly and plateaus at a lower match rate than TS–GRPO, whereas TS–GRPO reaches a higher final match on TS–SQL and maintains comparable execution on BIRD–dbt (Figures~\ref{fig:thinkquel_match}–~\ref{fig:gspo_match}; Table~\ref{tab:results}). 

The itmized reward plots for Thinkquel and GSPO over training steps can be found in Appendix~\ref{app:itemized-rewards}.

\textbf{Takeaways.} (i) Execution-aligned, span-aware optimization improves stability and match on TS\textendash SQL; (ii) explicit planning is not merely formatting---it provides measurable schema-grounding benefits; (iii) Relative to the SFT-only checkpoints, TS–GRPO adds a further +4.9 pp match on TS–SQL (53.6\% → 58.5\%) at 92.2\% execution (32B), and matches GSPO on BIRD–dbt (73.3\% vs 73.5\%) while keeping execution $\geq$ 92\%. Combined with the planning-aware SFT curriculum, this yields a compact 32B model that is execution-robust in-domain and competitive out-of-domain even compared to much larger proprietary models.

\section{Discussion and Future Work.}
Our results suggest that span-aware credit assignment is the principal driver of the stability gains we observe. TS-GRPO explicitly routes global, brittle signals (execution, result-match, plan-following) through a sequence-level update on the SQL span while keeping token-level updates for local, structural signals on the planning span; this design reduces variance, curbs cross-span credit leakage, and better matches the error surface of text-to-dbt generation than uniform token-only or sequence-only objectives.

Most residual failures stem from schema reference errors---the model occasionally names a non-existent table or column. These mistakes persist even after adding schema-linking rewards, indicating that out-of-vocabulary columns and ambiguous synonyms still slip through. This is consistent with long-standing evidence that accurate schema linking is a dominant challenge in text-to-SQL~\citep{wang2019ratsql,lin2020bridge,cao2021lgesql,gan2023reappraise}. We also observe a plan$\rightarrow$SQL drift (the final dbt deviates from the stated plan) and a small tail of execution-bounded failures (timeouts due to accidental Cartesian products or unselective joins). The negligible gain on BIRD-dbt compared to GSPO suggests that the TS-SQL complexity distribution may overweight long, intricate models; incorporating a broader mix of short/medium-complexity patterns should improve out-of-domain generalization. These limitations motivate the following directions:

\textbf{Wider Dataset Coverage; More Realistic Questions.}
We will rebalance TS-SQL toward a \emph{calibrated difficulty spectrum} (short/medium/long dbt models) and further diversify question styles beyond fully specified instructions to include realistic user questions and variation in vagueness and language style. 

\textbf{Integrate RL with Tool Use.}
To mitigate hallucination and improve schema grounding, we are moving toward a multi-turn RL setting where the policy can interleave generation with tool calls, e.g., schema inspection and dry-run with appropriate reward designs. This turns the problem into decision-making over both text and tools, improving credit assignment for retrieval/verification steps.

\textbf{Cross-Warehouse Portability.}
Because portability is a core motivation for dbt, we will extend evaluation across multiple warehouses (e.g., Snowflake, Databricks, Redshift, BigQuery) and inject dialect-sensitive cases into training. This stresses the model’s ability to keep logic invariant while adapting to warehouse-specific idioms.

\section{Conclusion}

We introduced Thinkquel, a system that includes a rigorous text-to-dbt synthetic data generation pipeline and a model that combines a two-stage SFT curriculum with Token–Sequence GRPO (TS–GRPO)—a span-aware RL objective that routes sequence-level updates to the SQL/dbt answer and token-level updates to the planning span. On Spider, TS–GRPO exhibits faster and smoother convergence of the execution-match reward than GRPO and GSPO under identical training conditions, consistent with the intended separation of credit assignment. On our TS–SQL benchmark, Thinkquel (32B) attains 92.2\% execution and 58.5\% match; on BIRD–dbt (1,482) it reaches 92.2\% execution and 72.3\% match and remains competitive at convergence—matching or slightly exceeding GSPO within the $\leq$ 32B class. These gains build on the planning-aware SFT curriculum, which contributes the majority of the jump from the base model. During reinforcement learning, TS–GRPO narrows the remaining gap after SFT and stabilizes training, while the explicit planning provides measurable schema grounding and reduces error propagation.

\bibliographystyle{apalike}
\bibliography{references}

\appendix
\section{System Prompt for Spider Dataset}
\label{app:system-prompt}
For our experiments on the Spider dataset, we use the following system prompt:
\begin{verbatim}
You are a helpful assistant specialized in generating SQL queries.

Requirements:
For the task given by the user, please think carefully 
and format your response as follows:
<think> 
{{your reasoning about how to solve the task}}
```yml 
{{source tables you plan to use}} 
```
</think>
<answer>
{{your final SQL query}} 
</answer>

Example Response:
<think>
To solve this problem, I need to:
1. Count the number of records in the head table
2. Filter by age > 56
3. Use COUNT(*) to get the total count
```yml
spider1_department_management.PUBLIC.head
```
</think>
<answer>
SELECT COUNT(*) FROM spider1_department_management.PUBLIC.head WHERE age > 56;
</answer>

{schema_text}
\end{verbatim}

\section{Itemized Rewards Dynamics for Thinkquel 32B Training vs. GSPO}
\label{app:itemized-rewards}

\begin{figure}[h!]
    \centering
    \includegraphics[width=1.0\textwidth]{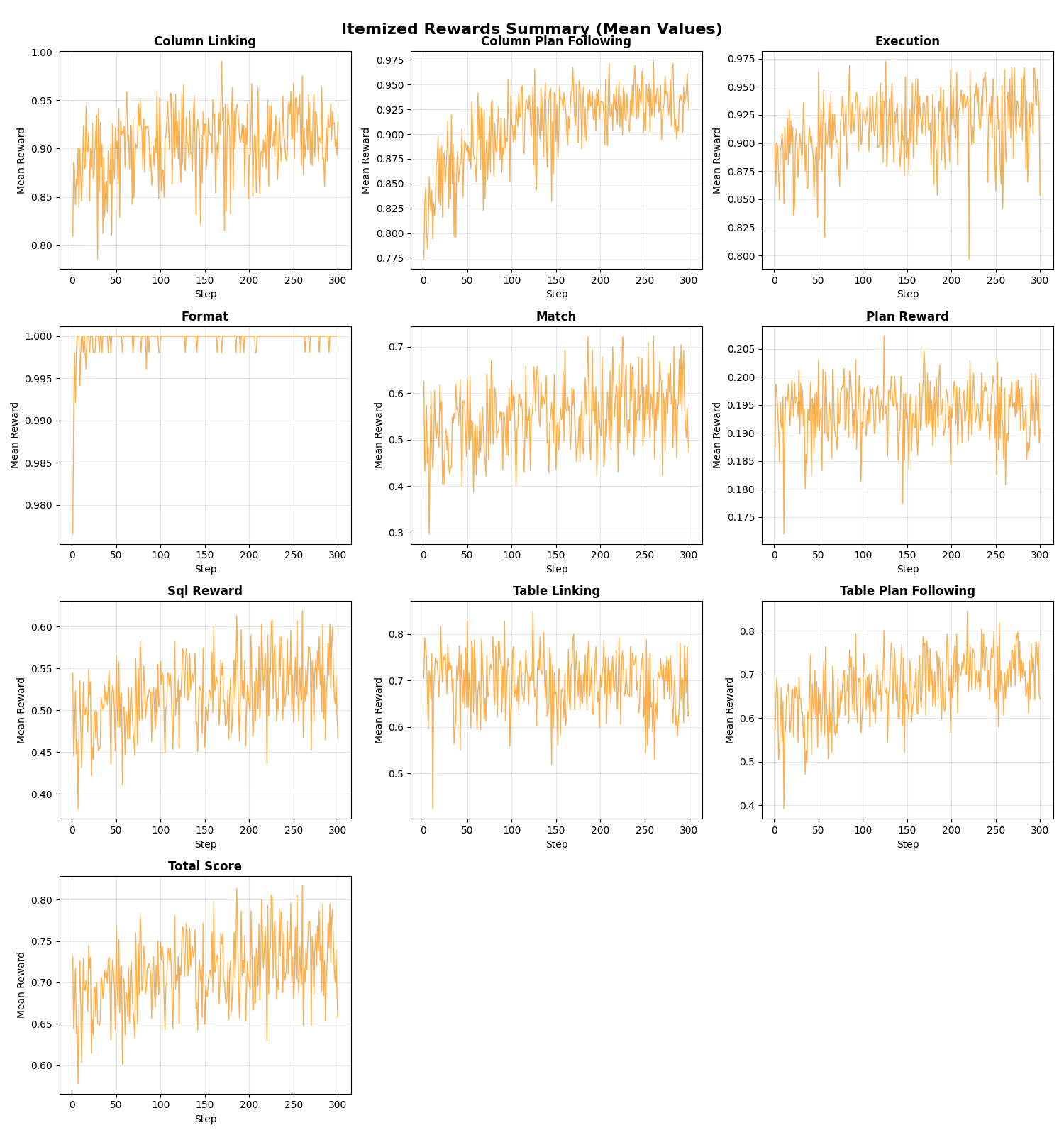}
    \caption{Learning Curves of Thinkquel on TS-SQL: All Itemized Rewards}
    \label{fig:thinkquel_all}
\end{figure}

\begin{figure}[h!]
    \centering
    \includegraphics[width=1.0\textwidth]{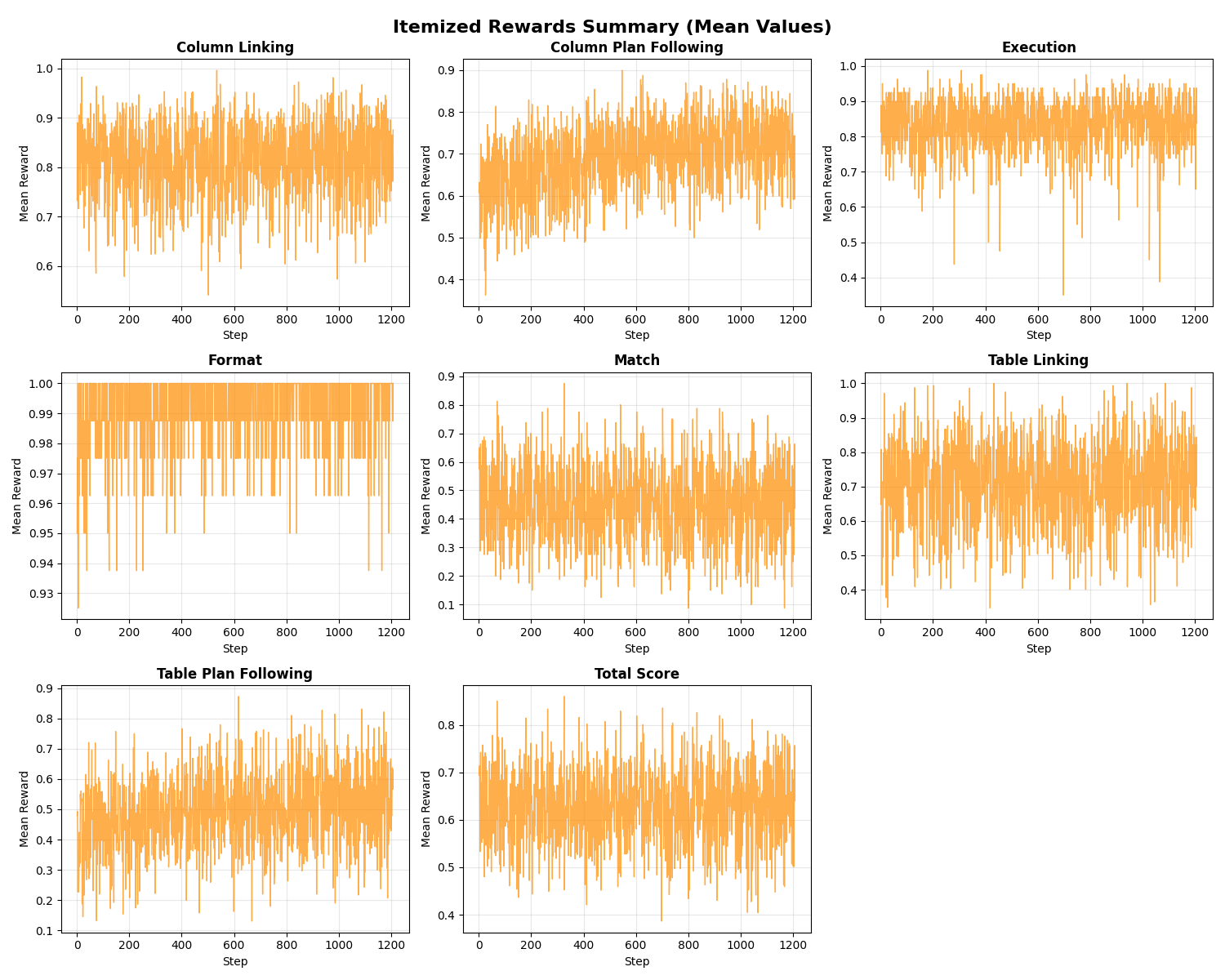}
    \caption{Learning Curves of GSPO on TS-SQL: All Itemized Rewards}
    \label{fig:gspo_all}
\end{figure}

\end{document}